\documentclass[runningheads]{llncs}
\usepackage{booktabs}
\usepackage{hyperref}
\usepackage{longtable}
\usepackage{array}
\usepackage{pifont} 
\usepackage{xcolor} 
\newcommand{\cmark}{{\raisebox{0pt}[0pt][0pt]{\color{green}\ding{51}}}} 
\newcommand{\xmark}{{\raisebox{0pt}[0pt][0pt]{\color{orange}\ding{55}}}} 
\newcommand{\pmark}{{\raisebox{0pt}[0pt][0pt]{\color{green}\ding{59}\ding{59}}}} 
\newcommand{\bmark}{{\raisebox{0pt}[0pt][0pt]{\color{green}\ding{59}}}} 

\usepackage{graphicx}
\begin{document}
\title{Callico: a Versatile Open-Source Document Image Annotation Platform}
%
%
\author{Christopher Kermorvant\orcidID{0000-0002-7508-4080}\and 
Eva Bardou \and
Manon Blanco \and Bastien Abadie
}
\authorrunning{C. Kermorvant et al.}
%
\institute{TEKLIA, Paris, France 
\email{kermorvant@teklia.com}\\
\url{https://teklia.com} }
\maketitle              
\begin{abstract}
This paper presents Callico, a  web-based open source platform designed to simplify the annotation process in document recognition projects. The move towards data-centric AI in machine learning and deep learning underscores the importance of high-quality data, and the need for specialised tools that increase the efficiency and effectiveness of generating such data. For document image annotation, Callico offers dual-display annotation for digitised documents, enabling simultaneous visualisation and annotation of scanned images and text. This capability is critical for OCR and HTR model training, document layout analysis, named entity recognition, form-based key value annotation or hierarchical structure annotation with element grouping. The platform supports collaborative annotation with versatile features backed by a commitment to open source development, high-quality code standards and easy deployment via Docker. Illustrative use cases - including the transcription of the Belfort municipal registers, the indexing of French World War II prisoners for the ICRC, and the extraction of personal information from the Socface project's census lists - demonstrate Callico's applicability and utility.

\end{abstract}
\section{Introduction}
In the field of document recognition, machine learning has become a fundamental component, forming the core of today's methods and applications. The advent of deep learning has further reinforced its dominance, marking a significant shift towards a unified modelling approach across different tasks. This convergence of models means that performance differences are increasingly attributed to the data - both its quantity and quality - rather than to the models themselves. This realisation has given rise to the data-centric AI movement\cite{zha2023}, which prioritises the systematic engineering of data over the pursuit of more sophisticated models within static datasets.

These developments highlight the importance of tools that enable the creation and refinement of annotations. These tools are critical to document recognition projects that demand not only efficiency and cost-effectiveness, but also the highest standards of data quality. Despite the availability of numerous annotation platforms, the specific requirements of document image recognition demand tools that are specifically designed to meet these complex needs. 

The principle that "more data beats a cleverer algorithm"\cite{domingos2012} carries significant weight in the field of machine and deep learning. It highlights a fundamental strategy in which increasing the size of the dataset often proves more beneficial and cost-effective than the laborious process of increasing algorithmic complexity. This approach is particularly relevant in scenarios where data scarcity poses a challenge to model performance. Expanding the volume of annotated data through efficient annotation processes can yield significant improvements, making it a preferred strategy over labour-intensive algorithmic optimisations. This underscores the critical need for annotation tools that can streamline the data collection and annotation process, allowing training datasets to be expanded in a resource-efficient manner without requiring excessive computational or specialised human resources.

However, the quantity of data is not the only determinant of a model's effectiveness; the quality of the data plays a crucial role, particularly in scenarios involving small to medium-sized data sets. In such cases, "no better data than good quality data" becomes a guiding maxim. The delicate balance between quality and quantity should be carefully managed in the design of annotation tools, ensuring that the annotated data not only contributes to the volume required for training, but also meets the highest quality standards. 

Efficiency in data annotation is paramount, given the potential cost and resource intensity of the process. This becomes even more critical in specialised domains where annotators, often highly skilled end-users, have limited time to devote to such tasks. To address this challenge, advanced annotation tools must be  designed to optimize the process and increase contributor productivity without compromising the quality of the annotated data. Moreover, effective management of the contributor team and their tasks is critical to the success of data annotation projects. Modern annotation tools must include sophisticated features for task assignment, progress tracking, quality control and validation.

The domain of document image annotation presents unique challenges that are not fully addressed by traditional image or text annotation tools. Document annotation tasks require to access both the visual elements of the document and its textual content. This dual requirement underscores the need for specialised tools since traditional image or text annotation tools are not able to  integrate these two critical aspects of document annotation. 

In this paper, our contribution is the presentation of Callico, a novel web-based open-source platform for collaborative document annotation. The features of Callico encompass the following key functionalities:
\begin{itemize}
    \item Dual display annotation for digitised documents: Callico  enables simultaneous visualisation and annotation of scanned images and text. This feature is particularly useful for training OCR and HTR models, analysing document layout and recognising named entities.

\item Collaborative  annotation: Callico supports collaborative efforts by allowing team members or volunteers to join and contribute to open or closed annotation campaigns.

\item Versatile annotation capabilities: the platform supports a wide range of tasks including text classification, manual transcription, layout annotation and information extraction. This versatility makes it a comprehensive solution for any type of annotation project.

\item Open source availability: the platform is released as open source software under the GNU AGPLv3 licence, ensuring that it is accessible to a wide audience and can be freely used, modified and distributed.

\item High quality, maintainable and evolvable code:  emphasis has been placed on software quality with the implementation of Continuous Integration/Continuous Deployment (CI/CD) practices to ensure that the code base remains maintainable, evolvable and high quality.

\item Easy on-premises deployment with Docker: the platform can be easily deployed on-premise using Docker, simplifying the installation process and enabling organisations to quickly set up and start annotating documents.
\end{itemize}

\section{Related work}
The landscape of document annotation tools is diverse, with numerous platforms offering specialised functionality designed for different facets of the annotation process. Despite this diversity, each platform has inherent limitations that can hinder the efficiency and comprehensiveness of document annotation projects.  In the following section, we examine the existing tools and identify the specific limitations that justify the proposal of such a new, more powerful platform.

Document transcription platforms have been essential to the field of handwritten text recognition since its inception \cite{tosselli2007,gatos14}. However, some of these platforms are no longer maintained \cite{garz2016,trivedi2019,seuret2018}, and others are not available as open-source, limiting their adaptability and accessibility \cite{vidalgorene2021}.

Transkribus \cite{kahle2017} is a platform that focuses on the transcription of historical handwritten and printed documents. It facilitates the training of custom handwriting recognition models (OCR/HTR) and provides a unified interface for transcription-related annotations. However, it lacks features for document classification or Named Entity Recognition training (NER) and the platform does not support team management or task assignment among annotators. Furthermore, it is worth noting that Transkribus is not open-source software, which limits the ability of the user community to modify, enhance, or integrate the platform to meet their specific needs. 

eScriptorium \cite{kiessling2019} is an open-source platform that provides tools for document segmentation and transcription, especially for historical documents. Although eScriptorium is open-source, its functionalities are largely confined to transcription tasks and do not extend to features such as document classification or named entity recognition, similar to Transkribus. Additionally, eScriptorium lacks a comprehensive system for task allocation and annotator management.

Pivan \cite{constum2023} is an open source, web-based platform designed to facilitate various stages of document processing, including layout analysis, transcription and named entity recognition (NER). It supports ALTO/METS file formats for input and output. The platform is built on Java/React technologies, provides an API and can be deployed on-premises via Docker. 

LabelStudio \cite{labelstudio2020} is a versatile annotation platform for various types of media. Although it includes an OCR mode for transcription, it is not specialised in document annotation.

Tagtog \cite{cejuela2014} specializes in annotating named entities in complete documents and supports PDFs with a text layer. However, it does not offer manual transcriptions or layout analysis.

Kili \cite{kili} is a platform for managing annotation campaigns across different types of media.  However, Kili, like Tagtog, is better suited for annotating modern electronic documents or PDFs, but lacks classification modes, metadata, and grouping.

Prodigy is a Python library used to create unit annotation tasks, primarily for NLP tasks. it does not support annotating documents with both text and images. 

FromThePage \cite{FromThePage} is a platform designed to simplify the transcription and annotation of historical documents, with an emphasis on collaboration and community engagement.  The platform supports full-text transcription interfaces with optional markup and annotation, configurable forms for indexing and structured data collection, and tools for document-level metadata description. FromThePage encourages collaboration through in-app discussions and "notes" sections on each page, where each revision is kept as a separate version, and enhances project management with features such as nightly emails to volunteers. Despite its comprehensive suite of tools for project configuration, promotion, management and multilingual support, a notable limitation of FromThePage is its lack of integration with machine learning technologies.

\begin{table}[ht]
\centering
\caption{Comparison of Document Annotation Platforms}
\label{tab:platform_comparison}
\begin{tabular}{
    >{\raggedright\arraybackslash}p{3cm}
    >{\centering\arraybackslash}p{2cm} 
    >{\centering\arraybackslash}p{2cm} 
    >{\centering\arraybackslash}p{1.5cm} 
    >{\centering\arraybackslash}p{1.5cm} 
    >{\centering\arraybackslash}p{1.5cm} 
}
\toprule
 & \textbf{Transkribus} & \textbf{eScript.} & \textbf{Pivan} & \textbf{Kili} & \textbf{Callico} \\
\midrule
Open-Source & \xmark & \cmark & \cmark & \xmark & \cmark \\
On-premise &  \xmark & \cmark & \cmark & \xmark & \cmark \\
Layout mode & \cmark & \cmark & \cmark & \cmark & \cmark \\
Text transcription & \cmark & \cmark & \cmark & \cmark & \cmark \\
Named-entity mode& \cmark & \xmark & \cmark & \cmark & \cmark \\
Key-Value mode & \xmark & \xmark & \xmark & \xmark & \cmark \\
Grouping mode & \xmark & \xmark & \xmark & \xmark & \cmark \\
User management & \cmark & \cmark & \cmark & \cmark & \cmark \\
Task management & \xmark & \xmark & \xmark & \cmark & \cmark \\
Quality management & \xmark & \xmark & \xmark & \pmark & \bmark \\
\bottomrule
\end{tabular}
\end{table}

\section{Design principles}
Callico's design principles focus on flexibility, user experience, and support for machine learning workflows. These principles guide the platform's architecture and feature set, ensuring that it meets the needs of both contributors and project managers in the creation, correction and validation of training data for machine learning and deep learning algorithms.

\paragraph{Simplified user interface:} A core principle of Callico is to optimise the user experience, which is achieved by adhering to the principle of "one task, one interface". This approach results in a lightweight UI that improves the contributor experience by minimising complexity. The main goal is to enable broad participation by keeping the contributor interface as simple as possible, while still providing project managers with the tools they need to track progress and manage campaigns effectively. Each type of annotation (e.g. layout, transcription, entities) corresponds to a specific annotation mode, allowing for targeted and efficient data collection and validation.

\paragraph{Designed for Machine Learning Data Preparation:} The primary function of Callico is to simplify the preparation of training data for machine learning and deep learning algorithms. This includes the creation, correction and validation of data to ensure its quality and relevance for algorithm training.

\paragraph{Multiple campaigns and annotation modes:} Projects in Callico can have multiple campaigns on the same elements, with each campaign potentially using a different annotation mode. This flexibility allows project managers to tailor the annotation process to the specific needs of each dataset or project phase.

\paragraph{Decentralised image hosting:} In line with its design for flexibility and scalability, Callico does not host images directly. Instead, images are retrieved from IIIF servers using the International Image Interoperability Framework to facilitate access and management of high quality digital images.

\paragraph{Open Source and Self-Hosting:} Callico is an open source project that emphasises transparency and community involvement in its development. It supports self-hosting, with deployment facilitated by Docker, allowing organisations to maintain control over their data and infrastructure. Detailed instructions for self-hosting can be found in the platform's documentation\footnote{https://doc.callico.eu/deploy/}.

\paragraph{Generic and extensible implementation:} Callico is built on a foundation of generic implementation principles that allow for easy extension and customisation. Key concepts such as providers (for retrieving images and data), campaign modes and export formats are designed to be generic, allowing users to extend or customise them according to their specific needs. This extensibility ensures that Callico can evolve with the changing needs of the document annotation community and the broader field of machine learning.

\section{Technical stack}
The technical implementation of Callico is based on a comprehensive stack that covers its technical foundation, deployment methodologies, and development practices.

Callico's core architecture is developed in Python, using the Django framework for the backend;  the user interface is built using vanilla JavaScript, bolstered by the Vue.JS framework. Callico uses PostgreSQL for database management, chosen for its strength and ability to handle complex data types and queries. Celery handles task queuing and execution, with Redis serving as the message broker, and Flower used for monitoring these tasks to ensure efficient background process management.

Storage solutions within Callico are addressed with MinIO, an S3-compatible high-performance object storage system that provides a reliable and scalable option for managing large volumes of data. MkDocs is used to facilitate documentation, which is a static site generator that simplifies the creation and maintenance of project documentation.

Callico deployment is streamlined through the use of Docker, complemented by Docker Compose for managing multi-container Docker applications. This setup simplifies the configuration and deployment process, making it straightforward and efficient. Traefik manages load balancing and request routing, serving as a modern reverse proxy and load balancer that enables service scalability and efficient request handling.

The platform supports integration with any SMTP provider for email services. It is also designed to be compatible with any S3-compatible storage provider and self-hosting through MinIO is also possible. Furthermore, hosting images requires a dependency on an IIIF server, which ensures high-quality image management and interaction capabilities.

On the development side, Git is used for version control to support collaborative development and efficient code management. Code quality and consistency are ensured through the use of pre-commit hooks. Comprehensive testing is achieved with Tox, which utilizes Pytest for backend tests and Selenium for end-to-end browser testing. Poedit simplifies localization and translation efforts, while Sentry handles application monitoring and error tracking. 

\section{Features of the platform}

\subsection{Project management}
Callico offers a range of features designed to optimise the creation and management of document annotation projects. Project creation is an exclusive feature reserved for staff users, ensuring that new projects are initiated with proper oversight. To meet a wide range of collaboration needs, projects can be designated as either private, where only managers can add contributors, or public, where anyone can sign up. The platform supports a member management system that categorises users into three roles: 'contributors' who are assigned annotation tasks, 'moderators' who monitor the quality of annotations and can answer questions from contributors, and 'managers' who configure projects, campaigns and task assignments. To facilitate the addition of new members, Callico provides a secure and easy-to-manage invitation link that can be invalidated or regenerated according to project requirements. 

\subsection{Campaign management}
At the start of a campaign, project managers need to select the annotation mode that best suits their project's objectives and specify parameters such as classes, element types, metadata types and entity types.

Central to setting up a campaign is the development of an annotation guide, which allows managers to provide clear instructions to annotators. This guide acts as an essential roadmap, ensuring that all participants understand the campaign's objectives and methodology. To encourage high quality annotation and to assess agreement between annotators, Callico allows double annotation to be enabled for a given sample.

The platform offers flexibility in task allocation by configuring batch sizes for claiming tasks, ranging from on-demand single tasks to pre-allocated batches. This allows for adaptive workflow management to accommodate the varying availability and workload preferences of contributors. During the annotation process, annotators are provided with image context for the target element, enhancing their ability to make informed annotations by situating the element within its broader visual framework.

A key component of campaign management is the ability to visualise overall progress. This feature gives project managers and contributors alike a clear overview of the campaign's progress, facilitating timely adjustments and motivation. Once the annotation effort is complete, the platform supports the export of annotations to Arkindex or in various formats, including CSV and XLSX, allowing the data to be incorporated into further analysis or model training.

\subsection{Annotation modes}
Callico supports a range of annotation modes for campaigns, each designed to address specific aspects of document analysis and processing.  Here is a description of the different annotation modes available:

\paragraph{Image Classification:} This mode allows users to categorise images or parts of images using predefined classes. It is particularly useful for sorting images into thematic or typological groups.

\begin{figure}[t]
\includegraphics[width=\textwidth]{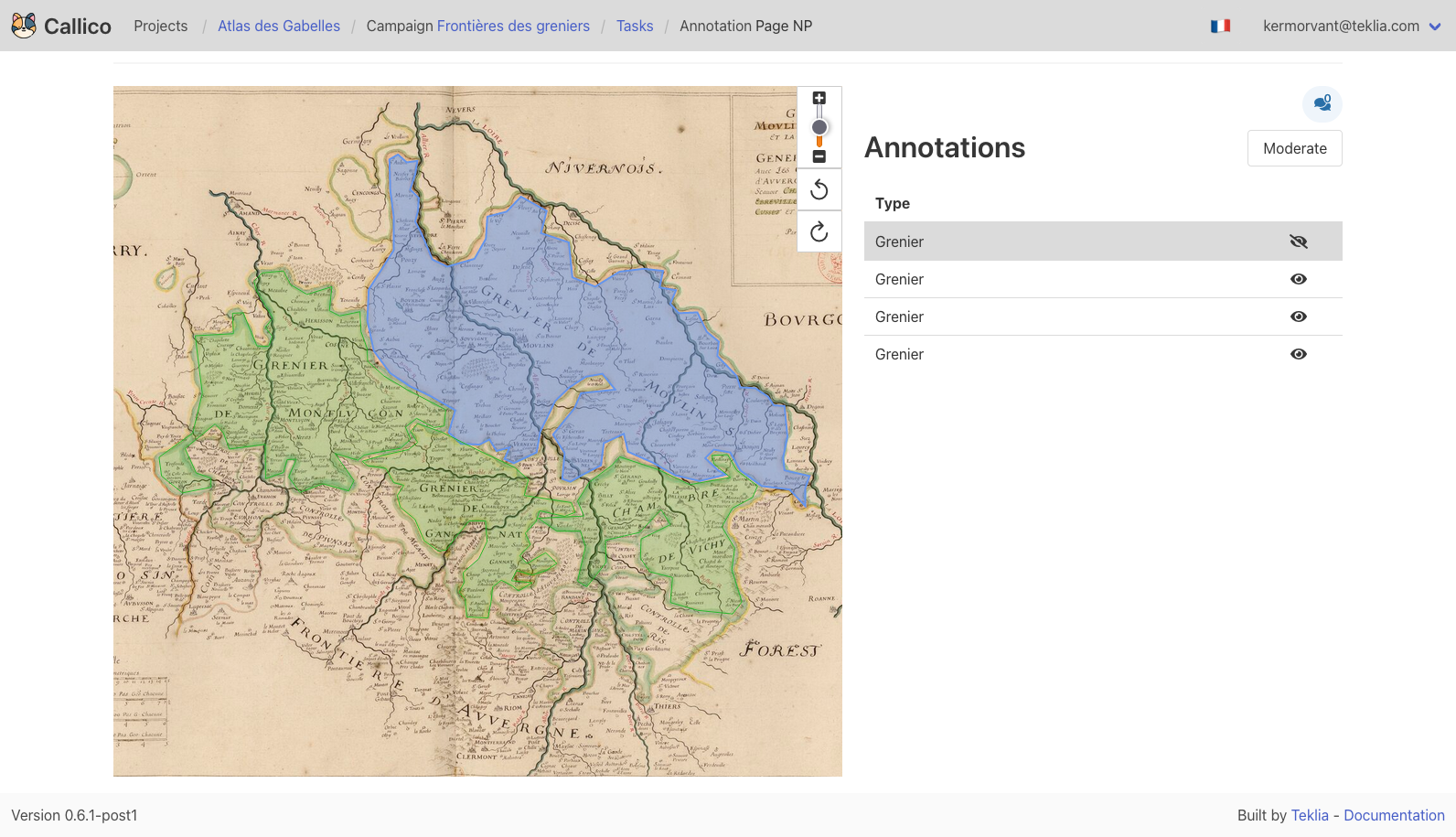}
\caption{Annotation mode for Document Structuring: an example of annotating different zones on a map using Document Structure mode, which allows the user to locate and type different zones on the document.} \label{zone}
\end{figure}

\paragraph{Document Structuring:} This  mode allows users to annotate the physical structure of documents by defining their constituent elements. This includes identifying and labelling different sections, paragraphs, headings and other structural components. An example of annotating different zones on a map using this mode is shown in the Figure \ref{zone}

\begin{figure}[t]
\includegraphics[width=\textwidth]{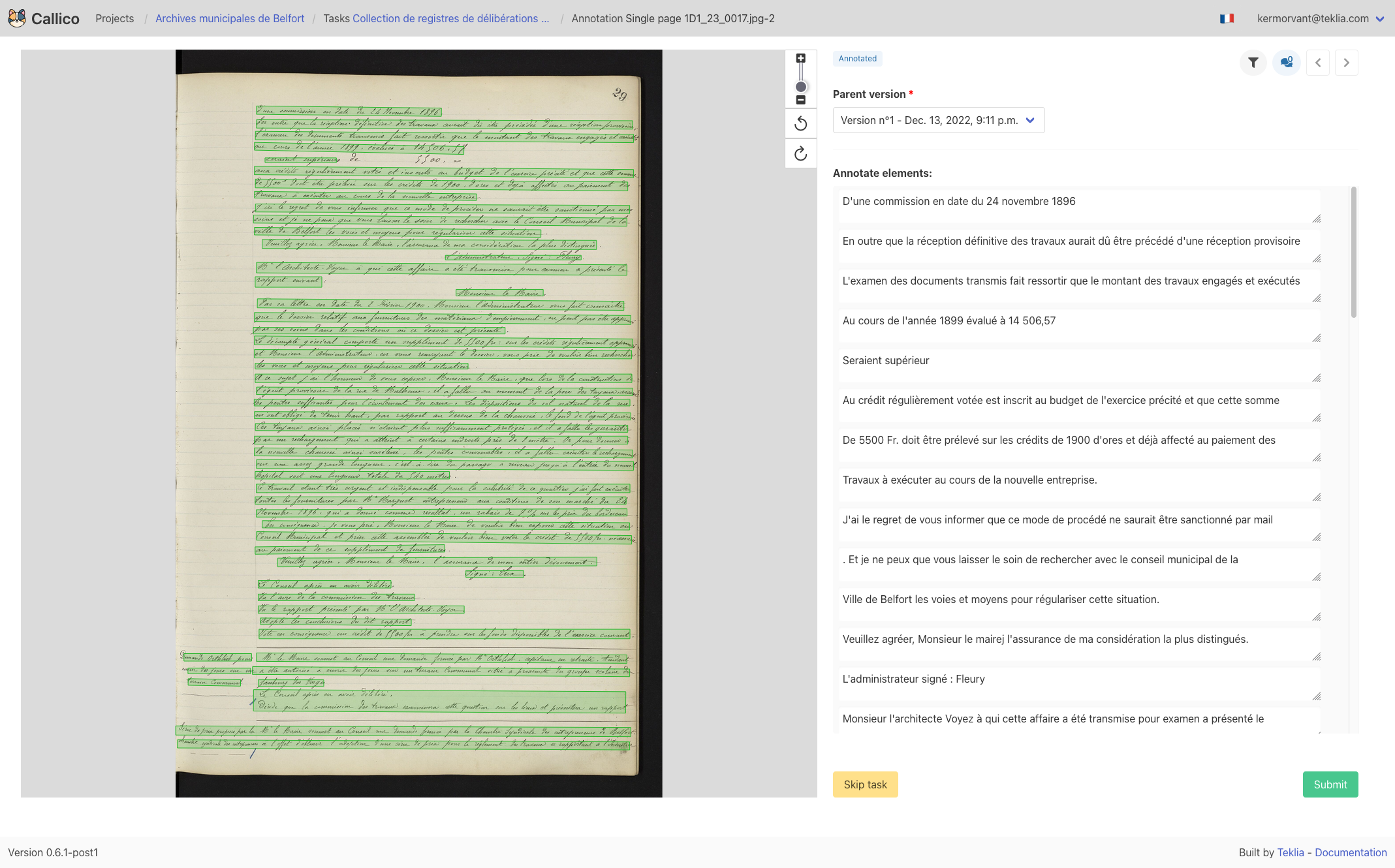}
\caption{Annotation mode for Text Transcription: an example of line transcription with image and text side by side.} \label{belfort}
\end{figure}

\paragraph{Text Transcription:} this mode is designed for manual transcription of printed or handwritten documents. It offers specialised options such as line-by-line or page-by-page views, facilitating accurate transcription by providing annotators with a focused and manageable segment of text at a time. An example text transcription using this mode is shown in the Figure \ref{belfort}

\begin{figure}[t]
\includegraphics[width=\textwidth]{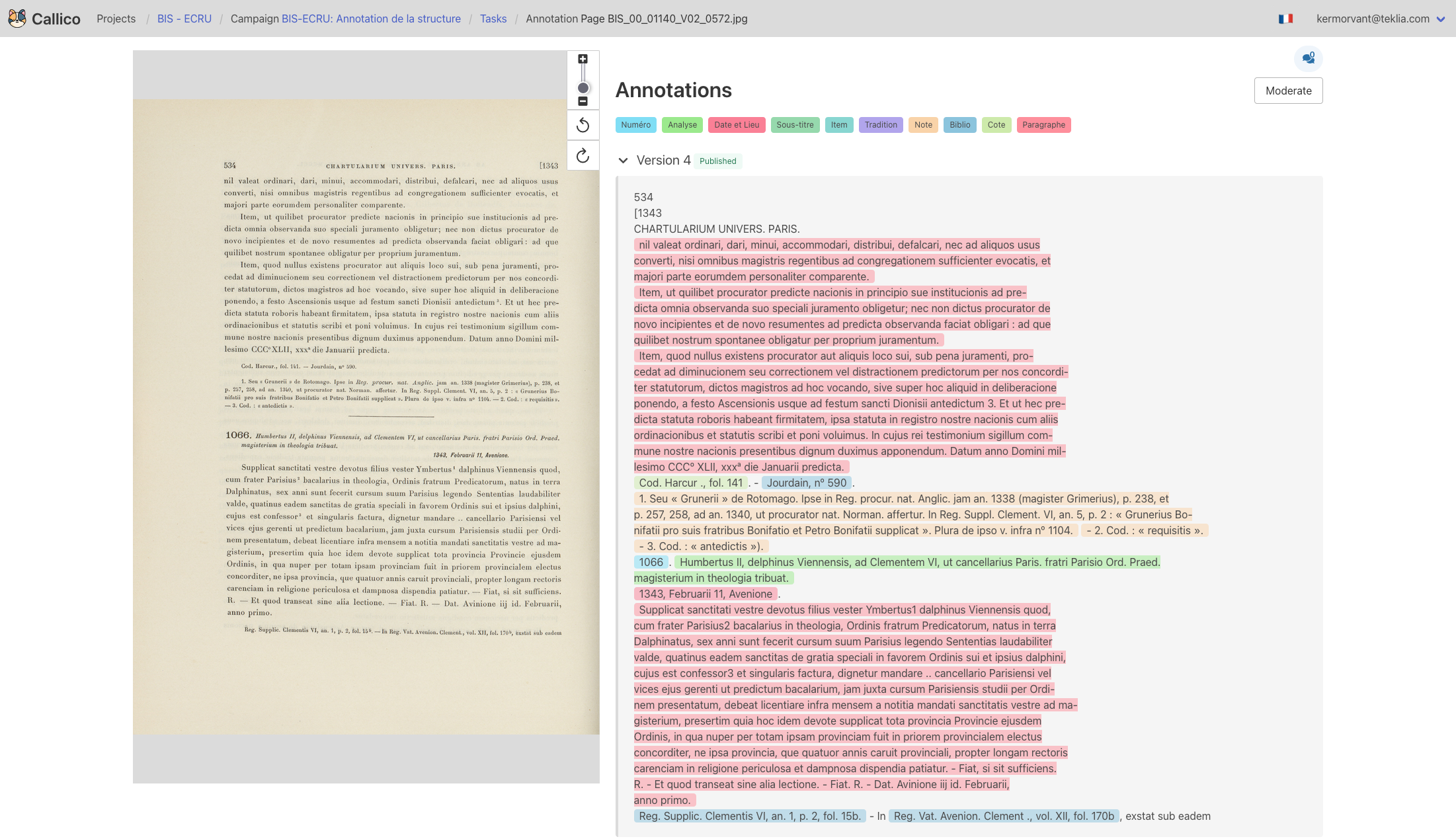}
\caption{Annotation mode for Named Entities: an example of annotating entities on a text by defining their range and type.} \label{ner}
\end{figure}

\paragraph{Named entity on text:} this mode focuses on the identification of named entities within the text, including their positions and types. It allows the tagging of proper names, places, organizations and other specific entities, adding a layer of semantic information to the textual data. An example of annotating entities on text is shown in the Figure \ref{ner}

\begin{figure}[t]
\includegraphics[width=\textwidth]{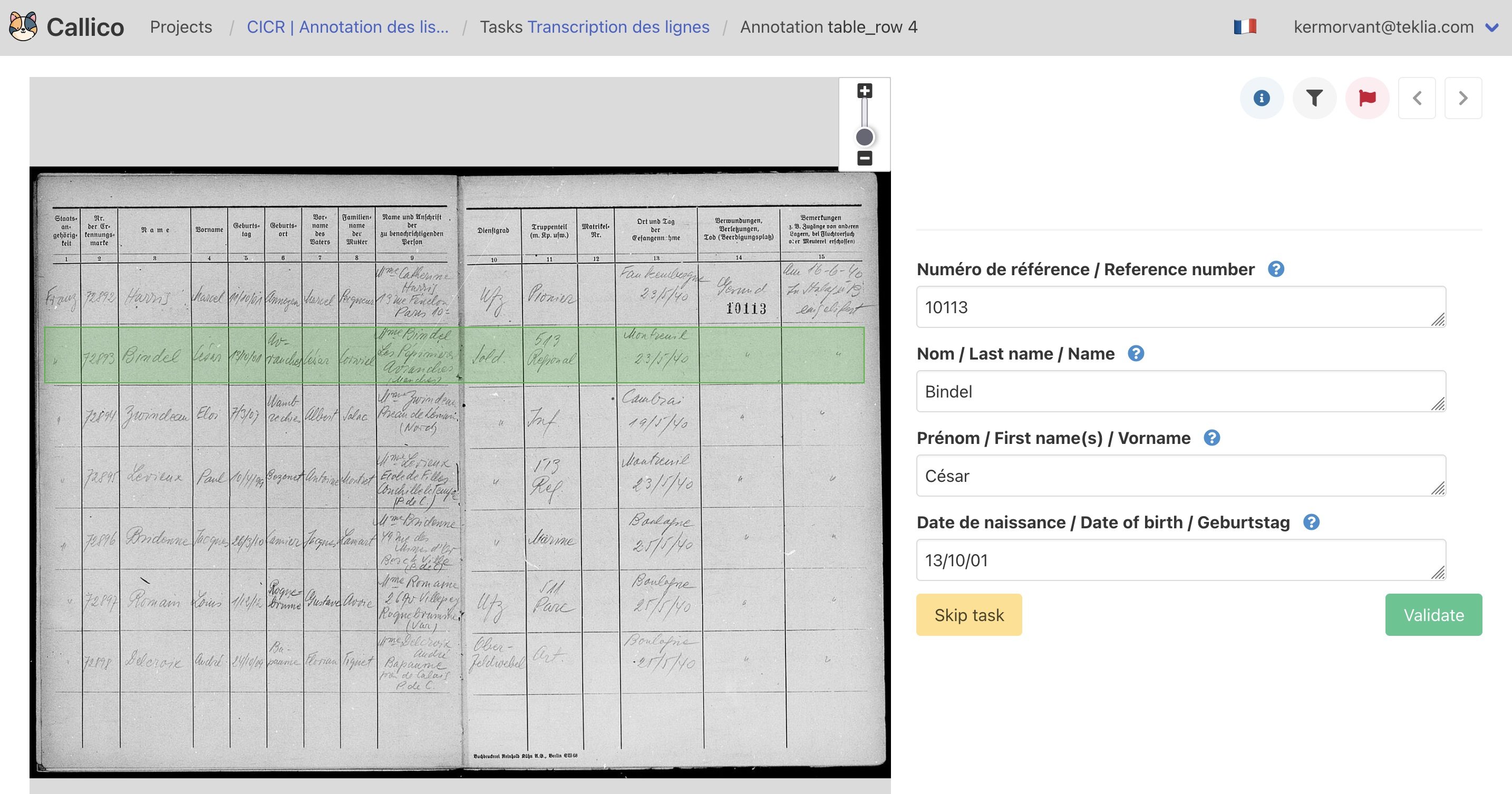}
\caption{Annotation mode for Key-Value Information: an example of annotating personal information from a table, with line highlighting.} \label{icrc}
\end{figure}

\paragraph{Key-Value Information:} this annotation mode allows users to define and extract metadata associated with document pages, such as topics or form-based information. This mode allows essential information to be identified without the need to directly associate entities with the text, providing a structured way of enriching the document's metadata. An example of personal information extraction using the key-value mode is shown in the Figure \ref{icrc}.

\begin{figure}[t]
\includegraphics[width=\textwidth]{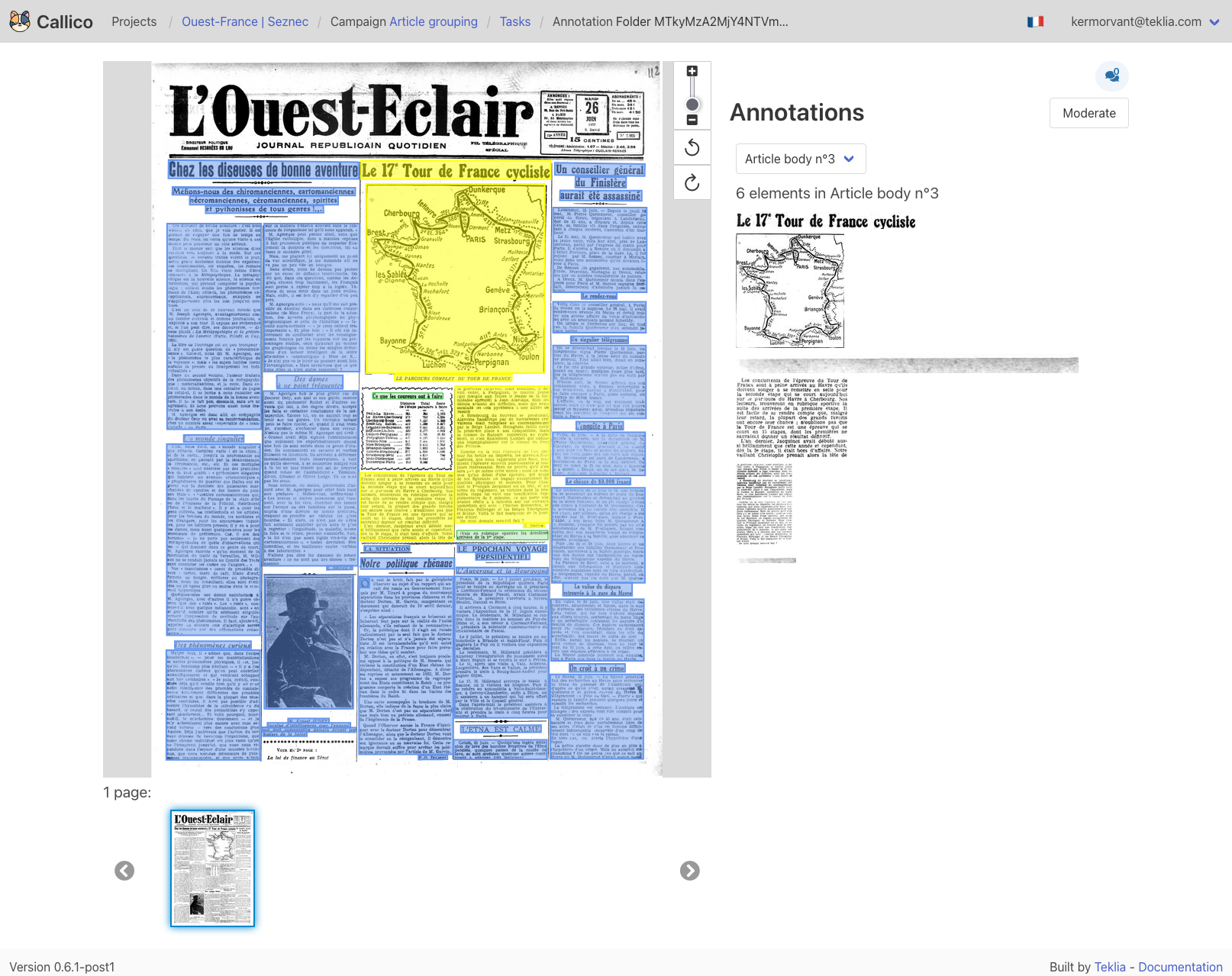}
\caption{Annotation mode for Element Grouping: an example of annotating a newspaper by grouping the different elements of each article.} \label{grouping}
\end{figure}

\paragraph{Element Grouping:} the element grouping mode allows users to group elements on a page to annotate a hierarchical structure. This is particularly useful for grouping different paragraphs of a deed in a registry or the different elements of a newspaper article (text, images, captions) to capture the relational and organizational aspects of the content. An example of newspaper annotation for grouping articles is shown in the Figure \ref{grouping}

\subsection{Task management}
Callico provides task management for document annotation projects with a system that simplifies the creation, distribution and monitoring of tasks. The platform allows managers to preview and refine tasks before making them available, ensuring that only well-defined tasks reach contributors. An efficient assignment system distributes tasks to volunteers either sequentially or randomly, effectively balancing the workload.

Tasks can be easily filtered by status (draft, pending, annotated, validated, rejected, skipped), user feedback (no feedback, commented, uncertain) or specific users, enabling targeted management and rapid response to requests. The annotation process is designed for ease of use, allowing contributors to claim tasks, skip inappropriate tasks and annotate with guidance from model predictions. Corrections to annotations are facilitated by easy submission of revised versions.

Moderators and managers have access to detailed visualisation tools for quality control and can moderate tasks to ensure accuracy. Unassigned tasks are returned to the available pool, ensuring that all tasks are addressed. In addition, a built-in commenting system supports active discussion among all project participants, encouraging collaboration and continuous improvement.

\section{Use cases}
In this section we present three different projects that have effectively used Callico to meet their specific needs for training data generation and prediction validation. Callico's versatility is illustrated by its application in the collaborative transcription of historical registers for the City of Belfort, the indexing of handwritten lists of French prisoners of war from the Second World War for the International Committee of the Red Cross, and the transcription and structuring of handwritten French census documents for the Socface project. 

\subsection{Transcription of registers}

The Municipal Archives of Belfort launched a pilot project to facilitate the automatic transcription of council minutes, with the aim of digitising and processing some 18,500 pages of historical documents dating from 1790 to 1946. The aim of the project was to make these documents available online, allowing citizens to engage with the history of their city by browsing and searching through council records, thereby increasing transparency and participation in local government.

To support this goal, Callico was used in transcription mode for an open collaborative annotation campaign\cite{tarride23}. In this setup, the document page was presented with all lines of text pre-detected by a text recognition model and ready for transcription by users. Callico's configuration for this project was designed to allow public access, with self-registration for volunteers who could select tasks on demand, encouraging broad community participation. The campaign included 616 images from registered sites and involved 150 contributors. 

\subsection{Information extraction in tables}

Using a hybrid approach combining collaborative transcription and artificial intelligence, the International Committee of the Red Cross (ICRC) has embarked on a project to improve the accessibility of its Tracing Archives, which contain 36 million names of prisoners of war (POWs) from World War II. Each year, these archives receive some 3,000 requests from former POWs and their descendants seeking to verify prisoner-of-war records or to trace imprisoned ancestors. The aim of making these archives searchable online was to significantly broaden the base of beneficiaries by facilitating access to this important historical data.

Organised by national service, with the French service including records of French nationals and foreigners who enlisted in the French uniform, the tracing archives are a treasure trove of detailed information. Each page lists up to 8 identities in a 15-column table, providing comprehensive data such as nationality, prisoner of war number, full name, date of birth, parental names, family address, military rank and unit, capture details, state of health and transfers.

The project began with a targeted double-key annotation campaign on a sample of 500 pages from these handwritten lists, undertaken by a select group of ICRC archivists. Unlike traditional annotation campaigns, which require line recognition, transcription and information tagging, the models developed by TEKLIA have greatly simplified this process. Annotations were made by simply filling in a form for each line in Callico, greatly simplifying and speeding up the task.

The initial phase of annotating 500 pages, each with an average of 8 lines, was completed in about 60 hours by 30 annotators, averaging about 30 seconds per line. A deep learning model for information extraction was then trained on this dataset and applied to a further 5000 pages. This second batch was manually validated or corrected within Callico, using an interface identical to the initial annotation phase, but with the model pre-filling the values. During this phase, 38851 lines of individual records were validated or corrected by 273 contributors, with the median validation time reduced to 13 seconds per line, demonstrating the efficiency and effectiveness of combining AI with human collaboration in processing historical archives. The second set of 5000 pages was then used to improve the deep learning information extraction mode, which achieved a CER of less than 3\% and an extraction F1 measure of more than 95\% for all fields. The remaining 700,000 images from the French prisoner archives were then automatically processed using this model. Following this success, the ICRC has installed Callico on their own infrastructure to validate the results of processing the 700,000 pages and to initiate new document recognition projects, leveraging the platform's capabilities for future indexing challenges.

\subsection{Personal information extraction and household grouping in census}
\begin{figure}[t]
\includegraphics[width=\textwidth]{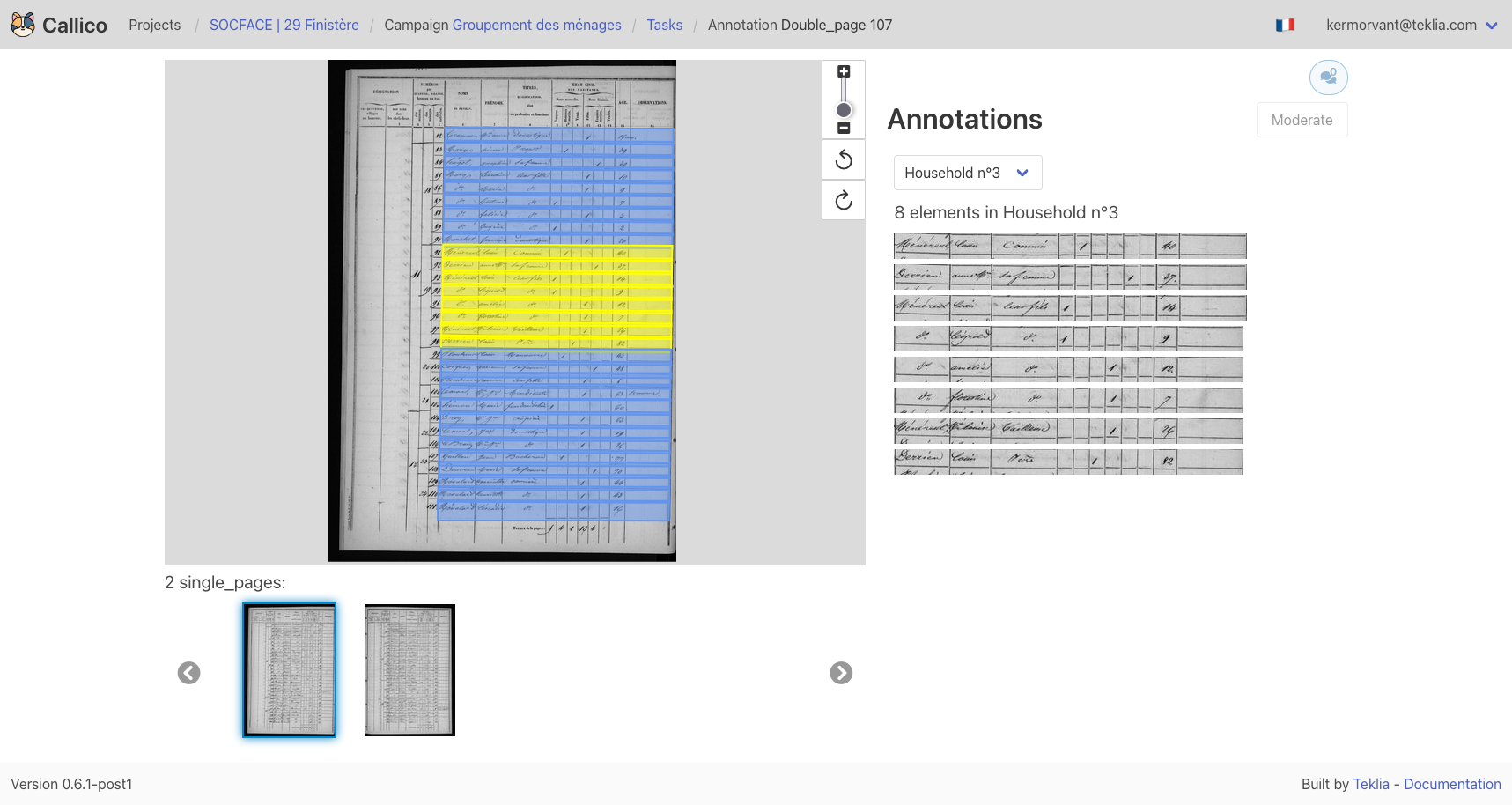}
\caption{} \label{socface}
\end{figure}
The Socface project is a collaborative effort between archivists, demographers and computer scientists to analyse and extract information from the French historical census on an unprecedented scale. Its main objective is to process all the handwritten nominal census lists from 1836 to 1936 using automatic handwriting recognition. These lists, produced every five years, are organised spatially by municipality, district, hamlet or street, and contain details of households and individuals, along with specific characteristics such as name, year of birth and occupation. The aim of the project is to use this archival material to create a database of individuals living in France over a century, facilitating the analysis of social change over this period. A major outcome of Socface will be the public availability of these nominal lists, providing open access to hundreds of millions of records.

To create the training material for deep learning models, 100 individual pages were randomly selected from the contributions of 11 departmental archives, representing the diversity of pages, image qualities and table templates from all the years studied. These pages were then uploaded to the Callico platform for manual transcription of the text in the table rows\cite{boillet2024}. The project used two different annotation modes within Callico: the Key-Value mode for annotating individual information, which presents users with a cropped image of the individual and a data entry form; and the ElementGroup mode for grouping individuals into households, which displays the full image with individual zones highlighted for users to group by household, as presented Figure \ref{socface}.

To date, 22 annotation campaigns have been successfully completed, two for each of the 11 selected departmental archives. These campaigns have provided valuable training data for the information extraction model. Specifically, 33,815 rows of tables have been manually annotated for individual information and 532 pages have been annotated for household grouping, with the efforts of 70 contributors. Most of the annotations were moderated, validated or corrected by experts where necessary, to ensure the accuracy and reliability of the data collected for this project.

\section{Conclusion}

Callico represents a significant advancement in the field of document recognition, providing a robust, open source platform designed to meet the complex requirements of annotating digitised documents. Through its innovative features such as dual display annotation, collaborative annotation capabilities and flexibility in handling different annotation tasks, Callico addresses the critical need for high quality data generation in machine learning and deep learning projects. The case studies presented in this paper highlight Callico's potential to transform document recognition tasks by improving efficiency, data quality and collaboration across domains.

Our future work aims to further refine Callico by focusing on two main areas: improving quality assessment mechanisms within the annotation process and developing more sophisticated validation modes. Improvements in quality assessment will explore inter-annotator agreement and sequential annotation modes, providing a more robust framework for ensuring data integrity. In addition, we plan to expand the platform's validation capabilities by incorporating different strategies that rely on prediction quality scores to optimise human validation efforts. By asking users to validate or correct machine predictions based on these scores, Callico will better integrate human expertise with automated processes, further improving the accuracy and reliability of annotated data.

As the document recognition community continues to evolve, there is a growing need for tools that can adapt to the changing landscape of data-centric AI. In this regard, Callico's open source nature serves as an invitation to the wider community to participate in its development and refinement. We encourage researchers, developers and practitioners to contribute to the Callico ecosystem by developing new types of annotation, extending its import and export capabilities, and innovating new quality metrics that can further enhance its utility and effectiveness.

\section{Acknowledgments}
The development of Callico was partially funded by the French National Research Agency (ANR) under the Socface project (ANR-21-CE38-0013), by the France Relance plan (ANR-21-PRRD-0010-01) and  by the Research Council of Norway through the 328598 IKTPLUSS HuginMunin project.

%
%
%
 \bibliographystyle{splncs04}
 \bibliography{icdar24_callico}

\begin{thebibliography}{10}
\providecommand{\url}[1]{\texttt{#1}}
\providecommand{\urlprefix}{URL }
\providecommand{\doi}[1]{https://doi.org/#1}

\bibitem{FromThePage}
Fromthepage. \url{https://fromthepage.com/}, accessed: 2024-02-11

\bibitem{kili}
Kili technology. \url{https://kili-technology.com/}, accessed: 2024-02-11

\bibitem{boillet2024}
Boillet, M., Tarride, S., Schneider, Y., Abadie, B., Kesztenbaum, L., Kermorvant, C.: The {S}ocface project: Large-scale collection, processing, and analysis of a century of {F}rench censuses. In: 2024 17th IAPR International Conference on Document Analysis and Recognition (ICDAR) (2024)

\bibitem{cejuela2014}
Cejuela, J.M., McQuilton, P., Ponting, L., Marygold, S.J., Stefancsik, R., Millburn, G.H., Rost, B., Consortium, F.: tagtog: interactive and text-mining-assisted annotation of gene mentions in plos full-text articles. Database (Oxford)  (2014)

\bibitem{constum2023}
Constum, T., Bebin, F., Tranouez, P., Paquet, T.: Pivan: A web-platform for document annotation. Archiving Conference  \textbf{20},  53--56 (06 2023). \doi{10.2352/issn.2168-3204.2023.20.1.10}

\bibitem{domingos2012}
Domingos, P.: A few useful things to know about machine learning. Commun. ACM  \textbf{55}(10),  78–87 (oct 2012). \doi{10.1145/2347736.2347755}

\bibitem{garz2016}
Garz, A., Seuret, M., Simistira, F., Fischer, A., Ingold, R.: Creating ground truth for historical manuscripts with document graphs and scribbling interaction. In: 2016 12th IAPR Workshop on Document Analysis Systems (DAS). pp. 126--131 (2016). \doi{10.1109/DAS.2016.29}

\bibitem{gatos14}
Gatos, B., Louloudis, G., Causer, T., Grint, K., Romero, V., Sánchez, J.A., Toselli, A.H., Vidal, E.: Ground-truth production in the transcriptorium project. In: 2014 11th IAPR International Workshop on Document Analysis Systems. pp. 237--241 (2014)

\bibitem{kahle2017}
Kahle, P., Colutto, S., Hackl, G., Mühlberger, G.: Transkribus - a service platform for transcription, recognition and retrieval of historical documents. In: 2017 14th IAPR International Conference on Document Analysis and Recognition (ICDAR). vol.~04, pp. 19--24 (2017). \doi{10.1109/ICDAR.2017.307}

\bibitem{kiessling2019}
Kiessling, B., Tissot, R., Stokes, P., Stökl Ben~Ezra, D.: e{S}criptorium: An open source platform for historical document analysis. In: 2019 International Conference on Document Analysis and Recognition Workshops (ICDARW). vol.~2, pp. 19--19 (2019). \doi{10.1109/ICDARW.2019.10032}

\bibitem{seuret2018}
Seuret, M., Bouillon, M., Simistira, F., Würsch, M., Liwicki, M., Ingold, R.: A semi-automatized modular annotation tool for ancient manuscript annotation. In: 2018 13th IAPR International Workshop on Document Analysis Systems (DAS). pp. 340--344 (2018). \doi{10.1109/DAS.2018.80}

\bibitem{tarride23}
Tarride, S., Faine, T., Boillet, M., Mouch\`{e}re, H., Kermorvant, C.: Handwritten text recognition from crowdsourced annotations. In: Proceedings of the 7th International Workshop on Historical Document Imaging and Processing. p. 1–6. HIP '23, Association for Computing Machinery, New York, NY, USA (2023)

\bibitem{labelstudio2020}
Tkachenko, M., Malyuk, M., Holmanyuk, A., Liubimov, N.: {Label Studio}: Data labeling software (2020), \url{https://github.com/heartexlabs/label-studio}, open source software available from https://github.com/heartexlabs/label-studio

\bibitem{tosselli2007}
Toselli, A., Romero, V., Rodriguez, L., Vidal, E.: Computer assisted transcription of handwritten text images. In: Ninth International Conference on Document Analysis and Recognition (ICDAR 2007). vol.~2, pp. 944--948 (2007). \doi{10.1109/ICDAR.2007.4377054}

\bibitem{trivedi2019}
Trivedi, A., Sarvadevabhatla, R.K.: Hindola: A unified cloud-based platform for annotation, visualization and machine learning-based layout analysis of historical manuscripts. In: 2019 International Conference on Document Analysis and Recognition Workshops (ICDARW). vol.~2, pp. 31--35 (2019). \doi{10.1109/ICDARW.2019.10035}

\bibitem{vidalgorene2021}
Vidal-Gor{\`e}ne, C., Dupin, B., Decours-Perez, A., Riccioli, T.: A modular and automated annotation platform for handwritings: Evaluation on under-resourced languages. In: Document Analysis and Recognition -- ICDAR 2021. pp. 507--522. Springer International Publishing (2021)

\bibitem{zha2023}
Zha, D., Lai, K.H., Yang, F., Zou, N., Gao, H., Hu, X.: Data-centric ai: Techniques and future perspectives. In: Proceedings of the 29th ACM SIGKDD Conference on Knowledge Discovery and Data Mining. p. 5839–5840. KDD '23, Association for Computing Machinery, New York, NY, USA (2023). \doi{10.1145/3580305.3599553}

\end{thebibliography}

\end{document}